\documentclass[conference]{IEEEtran}
\ifCLASSINFOpdf
\else
\fi

\usepackage{graphicx}
\usepackage{longtable}
\usepackage{url}
\usepackage[linesnumbered,ruled,vlined]{algorithm2e}
\begin{document}
%
\title{Thematic context vector association based on event uncertainty for Twitter}

\author{\IEEEauthorblockN{Vaibhav Khatavkar}
\IEEEauthorblockA{
College of Engineering, Pune}
\and
\IEEEauthorblockN{Swapnil Mane}
\IEEEauthorblockA{College of Engineering, Pune}
\and
\IEEEauthorblockN{Parag Kulkarni}
\IEEEauthorblockA{College of Engineering, Pune}}


%


\maketitle

\begin{abstract}
Keyword extraction is a crucial process in text mining. 
The extraction of keywords with respective contextual events in Twitter data is a big challenge. The challenging issues are mainly because of the informality in the language used. The use of misspelled words, acronyms, and ambiguous terms causes informality. 
The extraction of keywords with informal language in current systems is pattern based or event based. In this paper, contextual keywords are extracted using thematic events with the help of data association.  The thematic context for events is identified using the uncertainty principle in the proposed system. The thematic contexts are weighed with the help of vectors called thematic context vectors which signifies the event as certain or uncertain.   The system is tested on the Twitter COVID-19 dataset and proves to be effective. The system extracts event-specific thematic context vectors from the test dataset and ranks them.  The extracted thematic context vectors are used for the clustering of contextual thematic vectors which improves silhouette coefficient by 0.5\% than state of art methods namely TF and TF-IDF. The thematic context vector can be used in other applications like Cyberbullying, sarcasm detection, figurative language detection, etc.
\end{abstract}


%
\IEEEpeerreviewmaketitle

\section{Introduction}
For text data, many times natural language processing is performed to develop systems for named entity recognition, product ranking, sentence ranking, recommender systems, etc.  Data Mining approaches to extract data from social media content are discussed in \cite{1}.Hate speech detection using deep learning approach is done by Dylan Grosz and Patricia Conde-Cespedes in their work \cite{34}. The researchers pointed out that size, noise, and dynamism are important aspects of social media content. Wahab Khan et al. \cite{2} in their work presents natural language processing machine approaches used till now. The researchers also explained how POS tagging, word sense disambiguation, and discourse analysis can be used in solving the ambiguity while processing text. There are also some developments in natural language inference that try to infer meaning from the text data \cite{3}. 

Tremendous amount of data in textual format is available on the Internet via Social Media, Blogs etc. \cite{4}. "Daily 500 million tweets, 294 emails while 65 billion messages on whatsapp are sent. There are about 300 million users which are identified as new users on Social Media Platforms." \cite{5} There are various domains of text mining where this data is converted into useful information.  Topic modeling algorithms like LDA are applied on twitter data to extract topics \cite{6}. Another topic model for twitter data is developed by David and Martin in \cite{7}  which is based on Author Topic Modelling (ATM).  David Alvarez-Melis and Martin Saveski in \cite{7} develop a topic model for tweets and their recipes by using two topic modeling techniques namely LDA and Author topic model. ATM proved to be more effective than LDA but consumes more time. Also, data clustering will give similar data clusters of tweets \cite{8}, the twitter data can be classified into various categories \cite{9}. Namugera et al. in their work \cite{10} correlated  topics using positive and negative sentiments of the tweets. They also noted that the negative tweets play a vital role in information spread over twitter. Twitter as time series and event identification. 

Richard J. Povinille in his paper \cite{11} applied Time series data mining on temporal financial data to find the hidden patterns present in it. He defined an objective function to find trading edge over stocks in financial data. The system was tested on investment transactions of the stock market like the New York stock exchange.   Masoud Makrehchi et al. used events as market loss and gain in social media text and created marketing strategies \cite{12}.  Selene Yue Xu \cite{13} developed a forecasting system for stock price using Yahoo Finance and google trend data.They performed ARIMA time series analysis and also correlated news with events. They concluded that the news is influential while predicting stock prices. Nikan Chavoshi et al. \cite{14} identify abnormal user accounts present in twitter they use lag-sensitive hashing technique and a warping invariant correlation measure to identify and correlate abnormal users.The dynamic nature of twitter time series data is exploited with the help of Dynamic Heartbeat Graph for event detection by Saeed et al. \cite{33}.

\section{Related work}
Attempts have been made to map Twitter data as time series and analyze it. Salvatore et al. \cite{15} in their work did qualitative and quantitative analysis of social media content with twitter as a base example. They constructed a timely socio-economic indicator for social media.  Marta et al. \cite{16} in their work extracted social, economic, and commercial indicators using Twitter time series analysis.  Alshaabi et al. \cite{17} analyzed tweets which are related to medical, political, and COVID-related operations. The tweets were collected for 24 languages over a period of a year, i.e. from March 2019 to March 2020. They observed that the term ‘virus’ was at peak value in all languages in the month of January and February.  Cakit et al. \cite{18} investigated the use of supervised soft computing techniques, namely, fuzzy time series (FTS), artificial neural network (ANN)-based FTS, and adaptive neuro-fuzzy inference systems (ANFIS) for predicting the emotional states expressed in Twitter data. Gonzalo et al. in \cite{19} identified critical events using Bayesian networks. The critical events were identified during natural disasters and social movements. Bayesian networks were used to find the correlation between the words. The patterns in the tweets prove to be useful  for classification.  Malayak et al. in their work \cite{20} used Twitter time series for predicting oscar winners by analyzing tweets,  retweets, mentions, hashtags, and favorites on Twitter from 2015 to 2019. Logistic regression and time series decomposition was used to forecast Oscar winners. 

Pratap et al. \cite{21} detected using classified and clustered tweets and their context. The proposed system can be used for monitoring traffic in great areas like highways. Arapostathis and Karantzia \cite{22} use Volunteered Geographic Information (VGI) and Twitter data for disaster management. the system identified the fire events and tracked and classified them. Chen and Terejanu \cite{23} worked on  sub-event identification. In their approach, they mapped  an outlier detection problem using   Kalman Filter, Gaussian Process, and Probabilistic Principal Component Analysis. Behzadan et al. \cite{24} detected events in social networks namely Twitter for cyber security. The detected events were classified using Convolutional Neural Network (CNN).

Hossny et al. \cite{25} have used Naive Bayes classifiers  to predict future events on the basis of historical events and live events such as protests, present in tweets. Hasan et al. \cite{26} combined a random indexing-based term vector model with locality sensitive hashing that will help to cluster tweets related to various events within a fixed time. This will help to identify more trending, active, and meaningful news from data. Katragadda et al. \cite{27} found credible events by using topic evaluation on the pattern of conversation present in tweets. Kolchyna et al. \cite{28} worked on the automatic identification of the optimum event window for Twitter data time series. After event identification, they performed event clustering (cluster type) and quantification. Grindrod et al. \cite{29} in their work used event spikes on Twitter time series. Eric Lai et al. in their work \cite{30} model topics on Twitter data in the spatiotemporal time domain by building relations among them. Yadav et al. \cite{31} modeled tweets related to  IPL T20 with temporal time series for the identification of events. The box office revenue trends were analyzed and forecasting was made by Arias et al. in \cite{17} by using events in Twitter time series. 

Typically event-related data acquisition is done by using keywords as search patterns in the available data. Size data is a major problem.  Some researchers report success in event detection  with the use of large Twitter datasets. \cite{32} proposed the event identification method to retrieve more events that are related to specific content. This method monitored the events and rated them by their relationship to predefined events using the term frequency-inverse document frequency (TF-IDF). More related arising events are inserted as a filter model in real time. 

In the proposed method, the thematic context is built on the available data which can be small or large. The thematic context comprises keywords and their association with events. The events can be derived from the search patterns which are queries on the data.

\section{Proposed Methodology}
\begin{figure*}[http]
\centering
\includegraphics[width=0.7\textwidth]{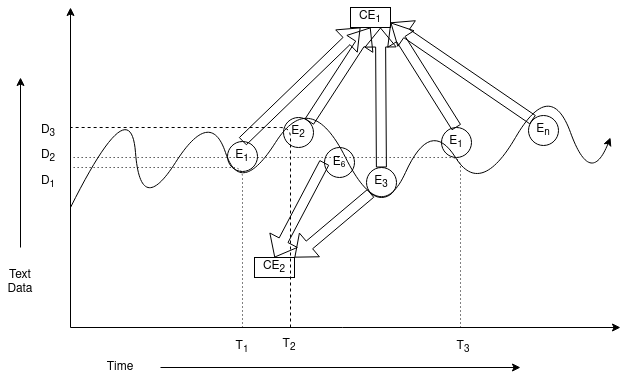}
\caption{Contextual Event Occurred at a time Y axis to be changed}
\label{Fig:propose}
\end{figure*}

The time series is usually a  tuple in the form of time and data i.e. TS = ${T_{i}, D_{i}}$. From the given time series, the events and data association can be interpreted for further processing. The time series data is shown in Fig \ref{Fig:propose}. The Figure shows various events $E_{1},E_{2}$, En which occur at various time instances. The events are keywords from the tweets  in the twitter time series represented as $D_{1},D_{2}…D_{n}$. 

The events in the time series  can be thematically associated to form thematic context vectors.  The uncertainty of  thematic vectors  for contextual events in the time series can be calculated and knowledge discovery can be done in order to rank the certain thematic context vectors in the time series data. 

\begin{figure*}[http]
\centering
\includegraphics[width=0.8\textwidth]{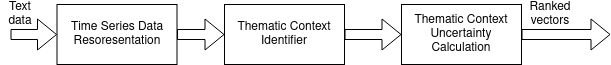}
\caption{The proposed system of thematic context vector association based on uncertainty}
\label{fig:proposeA}
\end{figure*}

The proposed system is shown in Fig \ref{fig:proposeA}.  The proposed system converts text data into time series data which is used for identifying thematic context. The thematic context is identified using data and event association.

The uncertainty for thematic context is derived using equation (1) given below. 

\begin{equation}
Uncertainty\ UN_{j}(i) =\ 1\ - IG_{j} (i) 
\end{equation}

\begin{equation}
Ranked Weight = UN_{j}(i) + Count(i) / Total count of all words
\end{equation}

where,
$IG_{j}(i) = C(j) * H_{j}(i)$ is Information gain for  in $j$ event, 

$H_{j}(i) = -(P(j)*log_{2}(1/P(j))+P(i)*log_{2}(1/P(i)) + P(i| j)*log_{2}P(i| j))$ is contextual entropy of event $j$ given word $i$,

$C(j)\ =\ count\ of\  event\ j\ in\ time\ series$.

After the uncertainty calculation  of the thematic context, its corresponding context vector is generated. The thematic context vector can be ranked and clustered. The algorithm to find thematic context is given below.  

\begin{algorithm}
\caption{Thematic ranking algorithm}

\textbf{Input}  Tweets and Query                                                                                   \\
\textbf{Outpt}  Ranked thematic context vectors with respective events                                             \\
\LinesNumbered

\While{$ t\ in\ tweets$}{ $ TS_{i} = t$}
/* Extract those events which are present in the dataset */ \\ 
$ E \leftarrow extract\_event(query) $ \\ $E_{k} \leftarrow extract\_keywords(TS_{i})$ 
\\ /*Define Ulist as list of Uncertainty for contextual events */ \\
\While{$e\ in\ E_{k}$}{ 
\While{$k\ in\ E_{k}$}{
\If{$cos(k,e)!=0$}{Derive $P(k|e)$ 
/*
append uncertainty of contextual event e with keyword k in Ulist */ \\                                       $Ulist.append( Un(k | e))$
} 
}
} 
/*Get thematic ranks of contextual events present in  Ulist along with the weights*/ \\ 
get\_rank(Ulist)                  
\end{algorithm}

The queries are the answers to be fetched from the Twitter data. The thematic context vectors for the queries will be generated. The ranked thematic context vectors will be in the form $CV_{j} = <{word_{1j}, uncertainty_{1j}}, {word_{2j}, uncertainty_{2j}… }$. The thematic context vector $CV_{j}$ will determine the theme of the event j.  

\section{Experimentation}

For experimentation  publicly available COVID-19 Twitter dataset is used as shown in Table \ref{table1}. By excluding all retweets and hashtags the total number of tweets is 3116 (Table \ref{table2}).

\begin{table}[http]
\centering
\caption{Details of Twitter dataset}
\resizebox{\linewidth}{!}{
\begin{tabular}{|l|c|l}
\cline{1-2}
\textbf{}                        & \multicolumn{1}{l|}{Dataset Details} &  \\ \cline{1-2}
Number of Tweets                 & 3116                                 &  \\ \cline{1-2}
Average words per tweets         & 10                                   &  \\ \cline{1-2}
Total number of words in dataset & 4729                                 &  \\ \cline{1-2}
Starting dateTime                & 2020-05-02 00:59                     &  \\ \cline{1-2}
Ending dateTime                  & 2020-05-02 00:03                     &  \\ \cline{1-2}
\end{tabular}

\label{table1}}
\end{table}

\begin{figure*}[http]
\includegraphics[width=\textwidth]{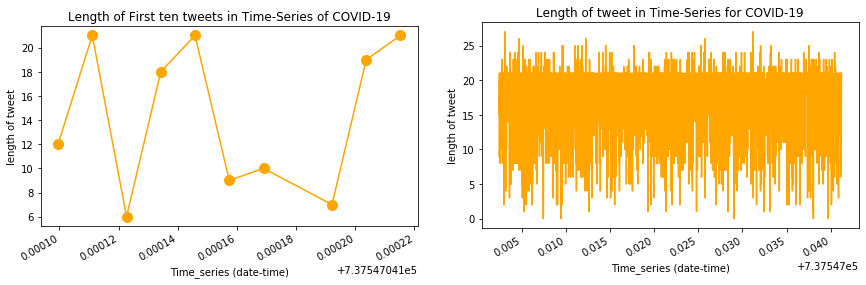}
\caption{Length of each tweet in time-series for COVID-19}
\label{fig:time1}
\end{figure*}

The tweets in the dataset were represented in time series as TS = {$T{i}, D{i}$} where T = {2020-05-02 00:59,2020-05-02 00:03}, D = keywords in tweets.  Fig \ref{fig:time1} shows  the  time series in which  the X axis is tweet time  and the Y axis is the number of keywords at that time. 

\begin{figure*}[http]
\includegraphics[width=\textwidth]{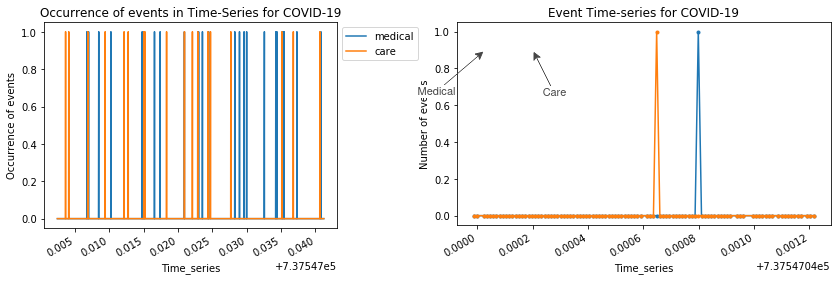}
\caption{Occurrence of each event in Time-Series for COVID-19}
\label{fig:time2}
\end{figure*}

Fig \ref{fig:time2} shows the event time series in which  the X axis is the time at which the tweets were made, and the Y axis is the number of events occurring at that time.  The number of events varies according to time. The event time series for this data holds a peak at value 1 which indicates the presence of the event. The continuous  consecutive  occurrence of events is depicted by a thick  line. Table \ref{fig:time2} shows the top 10 certain thematic context vectors for various events in the COVID-19 dataset

\par
From the event time series and keyword time series, the task to find thematic context between them. For thematic context vector identification, various queries were considered.   The thematic contextual keywords associated with the event were found and ranked.   

\begin{figure}[http]
\centering
\includegraphics[width=0.5\textwidth]{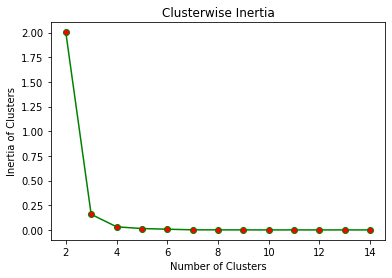}
\caption{Cluster-wise Inertia }
\label{fig:kmean}
\end{figure}

The K-means cluster is formed to get thematic events. In order to choose the number of clusters, the elbow method is used. It gives inertia per cluster. Fig \ref{fig:kmean} shows cluster-wise inertia with an elbow near five clusters for thematic event clustering. 

\par
\section{Result discussion}
For the tweets, queries were considered for extracting thematic context vectors.  For example,  the sample query 'What has been published about medical care?' was considered. In this query, “Medical” and “care”  are events. The contextual information will be derived using associated words in the tweets. The following table gives the keywords identified and their uncertainty values  for the above two events. The events are classified into two categories: certain and uncertain. Certain events will be identified with the help of keywords with an uncertainty value almost near 0. 

\begin{table*}[http]
\centering 

\caption{Identified  keywords and their uncertainty value for the two events}
\resizebox{\linewidth}{!}{
\begin{tabular}{|l|l|l|l|l|l|l|l|}
\hline
\multicolumn{4}{|c|}{\textbf{Certain Event}}         & \multicolumn{4}{c|}{\textbf{Uncertain Events}}         \\ \hline
\multicolumn{2}{|c|}{\textbf{Medical}} & \multicolumn{2}{c|}{\textbf{Care}} & \multicolumn{2}{c|}{\textbf{Medical}} & \multicolumn{2}{c|}{\textbf{Care}} \\ \hline
Keyword    & Rank Weight & Keyword     & Rank Weight & Keyword    & Rank Weight & Keyword       & Rank Weight \\ \hline
virus      & 0.007       & accidental  & 0.000523    & generally  & 1.327308    & providers     & 1.321089    \\ \hline
emergency  & 0.004       & pandemic    & 0.029304    & anymore    & 1.32914     & service       & 1.322659    \\ \hline
lockdown   & 0.005       & panic       & 0.000785    & panic      & 1.32757     & approximately & 1.321089    \\ \hline
medicine   & 0.001       & medical     & 0.006018    & government & 1.246714    & rest          & 1.237301    \\ \hline
nursing    & 0.001       & researchers & 0.002093    & private    & 1.238603    & neglect       & 1.321089    \\ \hline
facilities & 0.003       & news        & 0.014129    & political  & 1.12092     & heal          & 1.235993    \\ \hline
\end{tabular}
}
\label{table2}
\end{table*}

The thematic contextual vector for events ‘medical’  and ‘care’ are : 
\begin{itemize}
    \item CV$_{1}$ = \{(virus,0.007), (emergency,0.004), (lockdown,0.005), (medicine,0.001), (nursing,0.001), (facilities,0.003)\}

    \item CV$_{2}$ = \{(generally, 1.327308), (anymore, 1.32914), (panic, 1.32757), (government, 1.246714), (private,1.238603), (political, 1.12092)\}
\end{itemize}
The thematic context vector CV$_{1}$ certainly defines the context of the event medical than CV$_{2}$. Similarly, for event care, we get : 

\begin{itemize}
    \item CV$_{3}$ = \{(accidental,0.000523), (pandemic,0.029304), (panic,0.000785), (news,0.006018), (medical,0.002093), (researchers,0.014129)\}

    \item CV$_{4}$ =  \{(providers, 1.321089), (service, 1.322659), (approximately,1.321089), (rest, 1.237301), (neglect, 1.321089), (heal, 1.235993)\}
    
\end{itemize}    
The thematic context vector CV$_{3}$ will certainly define the context of the event care than CV$_{4}$.
Table \ref{table3} shows the top 10 certain thematic context vectors for various events in the COVID-19 dataset.

\begin{table*}[http]
\centering
\caption{Top 10 certain thematic context vectors for various events}
\label{table3}
\resizebox{\textwidth}{!}{%
\begin{tabular}{|c|cccccccccc|}
\hline
 &
  \multicolumn{10}{c|}{\textbf{Uncertain Events}} \\ \hline
\textbf{Rank} &
  \multicolumn{1}{c|}{\textbf{brilliant}} &
  \multicolumn{1}{c|}{\textbf{scientists}} &
  \multicolumn{1}{c|}{\textbf{announce}} &
  \multicolumn{1}{c|}{\textbf{medical}} &
  \multicolumn{1}{c|}{\textbf{care}} &
  \multicolumn{1}{c|}{\textbf{alive}} &
  \multicolumn{1}{c|}{\textbf{action}} &
  \multicolumn{1}{c|}{\textbf{covid}} &
  \multicolumn{1}{c|}{\textbf{corona}} &
  \textbf{\begin{tabular}[c]{@{}c@{}}corona-\\ virus\end{tabular}} \\ \hline
\textbf{0} &
  \multicolumn{1}{c|}{\begin{tabular}[c]{@{}c@{}}announ-\\ cement\end{tabular}} &
  \multicolumn{1}{c|}{korean} &
  \multicolumn{1}{c|}{announced} &
  \multicolumn{1}{c|}{virus} &
  \multicolumn{1}{c|}{accidental} &
  \multicolumn{1}{c|}{ask} &
  \multicolumn{1}{c|}{house} &
  \multicolumn{1}{c|}{covid} &
  \multicolumn{1}{c|}{futures} &
  gone \\ \hline
\textbf{1} &
  \multicolumn{1}{c|}{long} &
  \multicolumn{1}{c|}{people} &
  \multicolumn{1}{c|}{announcement} &
  \multicolumn{1}{c|}{emergency} &
  \multicolumn{1}{c|}{pandemic} &
  \multicolumn{1}{c|}{time} &
  \multicolumn{1}{c|}{trump} &
  \multicolumn{1}{c|}{black} &
  \multicolumn{1}{c|}{artist} &
  excuse \\ \hline
\textbf{2} &
  \multicolumn{1}{c|}{thought} &
  \multicolumn{1}{c|}{conclude} &
  \multicolumn{1}{c|}{heres} &
  \multicolumn{1}{c|}{lockdown} &
  \multicolumn{1}{c|}{panic} &
  \multicolumn{1}{c|}{decision} &
  \multicolumn{1}{c|}{amp} &
  \multicolumn{1}{c|}{blocking} &
  \multicolumn{1}{c|}{expecting} &
  six \\ \hline
\textbf{3} &
  \multicolumn{1}{c|}{tackle} &
  \multicolumn{1}{c|}{infected} &
  \multicolumn{1}{c|}{safeguard} &
  \multicolumn{1}{c|}{medicine} &
  \multicolumn{1}{c|}{medical} &
  \multicolumn{1}{c|}{president} &
  \multicolumn{1}{c|}{taken} &
  \multicolumn{1}{c|}{blocked} &
  \multicolumn{1}{c|}{excuse} &
  california \\ \hline
\textbf{4} &
  \multicolumn{1}{c|}{scientists} &
  \multicolumn{1}{c|}{cannot} &
  \multicolumn{1}{c|}{reopening} &
  \multicolumn{1}{c|}{nursing} &
  \multicolumn{1}{c|}{researchers} &
  \multicolumn{1}{c|}{guys} &
  \multicolumn{1}{c|}{decision} &
  \multicolumn{1}{c|}{block} &
  \multicolumn{1}{c|}{evidence} &
  globe \\ \hline
\textbf{5} &
  \multicolumn{1}{c|}{renounced} &
  \multicolumn{1}{c|}{south} &
  \multicolumn{1}{c|}{pleased} &
  \multicolumn{1}{c|}{die} &
  \multicolumn{1}{c|}{service} &
  \multicolumn{1}{c|}{omg} &
  \multicolumn{1}{c|}{deliberately} &
  \multicolumn{1}{c|}{bleach} &
  \multicolumn{1}{c|}{served} &
  birthday \\ \hline
\textbf{6} &
  \multicolumn{1}{c|}{national} &
  \multicolumn{1}{c|}{twice} &
  \multicolumn{1}{c|}{decision} &
  \multicolumn{1}{c|}{dean} &
  \multicolumn{1}{c|}{see} &
  \multicolumn{1}{c|}{mother} &
  \multicolumn{1}{c|}{new} &
  \multicolumn{1}{c|}{blaming} &
  \multicolumn{1}{c|}{set} &
  evidence \\ \hline
\textbf{7} &
  \multicolumn{1}{c|}{made} &
  \multicolumn{1}{c|}{outbreak} &
  \multicolumn{1}{c|}{may} &
  \multicolumn{1}{c|}{panic} &
  \multicolumn{1}{c|}{securely} &
  \multicolumn{1}{c|}{mercy} &
  \multicolumn{1}{c|}{designed} &
  \multicolumn{1}{c|}{blackhorse} &
  \multicolumn{1}{c|}{globe} &
  bad \\ \hline
\textbf{8} &
  \multicolumn{1}{c|}{today} &
  \multicolumn{1}{c|}{locally} &
  \multicolumn{1}{c|}{launched} &
  \multicolumn{1}{c|}{generally} &
  \multicolumn{1}{c|}{saving} &
  \multicolumn{1}{c|}{jose} &
  \multicolumn{1}{c|}{seem} &
  \multicolumn{1}{c|}{bitch} &
  \multicolumn{1}{c|}{espiritu} &
  nice \\ \hline
\textbf{9} &
  \multicolumn{1}{c|}{initiative} &
  \multicolumn{1}{c|}{caused} &
  \multicolumn{1}{c|}{latest} &
  \multicolumn{1}{c|}{saying} &
  \multicolumn{1}{c|}{jumped} &
  \multicolumn{1}{c|}{jones} &
  \multicolumn{1}{c|}{republicans} &
  \multicolumn{1}{c|}{brewery} &
  \multicolumn{1}{c|}{eric} &
  blaming \\ \hline
\end{tabular}%
}
\end{table*}

\begin{figure}[http]
\centering
\includegraphics[width=0.5\textwidth]{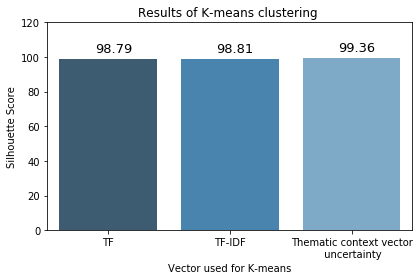}
\label{fig:res}
\caption{Results of K-means clustering }
\end{figure}
The results of clustering with TF, TF-IDF, and thematic context vectors are shown in Fig \ref{fig:res}.  The clustering with thematic context vector shows  an average 0.5\% improvement over both states of art methods.

\section{Conclusion}
Essential knowledge in the times series of Twitter data is a set of keywords for contextual events. Thematic context vectors are derived from the Twitter data which gives the context of the event. The queries are used to find the context of the event. The uncertainty principle is used to rank the thematic vectors in the events. The thematic context vector shows an uncertainty value of 0 for certain events in the data which are associated with each other. 
When event clustering is performed using the proposed method, the average improvement in silhouette coefficient is achieved by 0.5\% over clustering performed with TF and TF-IDF methods.  For contextual relation, the conditional probability of keywords with respective events is used which proves to be effective. This can be applied to generate automated events in the tweet  In applications  like cyberbullying, sarcasm detection, language models, etc. the thematic context vectors identified using the proposed system can be applied.



%

\bibliographystyle{plain}  	  
\bibliography{refer} 




\end{document}